\documentclass[10pt,twocolumn,letterpaper]{article}

\usepackage[pagenumbers]{cvpr} %

\usepackage{graphicx}
\usepackage{amsmath}
\usepackage{amssymb}
\usepackage{booktabs}
\usepackage{gensymb}
\usepackage{multirow}
\usepackage{hhline}
\usepackage{colortbl}
\usepackage{scalerel}
\usepackage{booktabs}
\usepackage{tabularx}
\usepackage{makecell}
\usepackage[normalem]{ulem}
\usepackage{inconsolata}
\usepackage[accsupp]{axessibility}
\usepackage[pagebackref, pageanchor=true, plainpages=false, pdfpagelabels, bookmarks, bookmarksnumbered]{hyperref}

\usepackage{enumitem} %
\usepackage{overpic} %
\usepackage[dvipsnames]{xcolor}
\usepackage{graphicx}

\setlist{itemjoin ={,\enspace},itemjoin* = { and\enspace}}
\definecolor{citecolor}{HTML}{0071bc}
\hypersetup{
  breaklinks   = true,
  colorlinks   = true, %
  urlcolor     = RubineRed, %
  linkcolor    = RubineRed, %
  citecolor    = citecolor %
}

\newcommand\blfootnote[1]{%
  \begingroup
  \renewcommand\thefootnote{}\footnote{#1}%
  \addtocounter{footnote}{-1}%
  \endgroup
}

\definecolor{turquoise}{cmyk}{0.65,0,0.1,0.3}
\definecolor{purple}{rgb}{0.65,0,0.65}
\definecolor{dark_green}{rgb}{0, 0.5, 0}
\definecolor{orange}{rgb}{0.8, 0.6, 0.2}
\definecolor{red}{rgb}{0.8, 0.2, 0.2}
\definecolor{darkred}{rgb}{0.6, 0.1, 0.05}
\definecolor{blueish}{rgb}{0.0, 0.3, .6}
\definecolor{light_gray}{rgb}{0.7, 0.7, .7}
\definecolor{pink}{rgb}{1, 0, 1}
\definecolor{greyblue}{rgb}{0.25, 0.25, 1}

\usepackage{blindtext}

\renewcommand{\paragraph}[1]{\vspace{0.5em}\noindent\textbf{#1}.}

\usepackage[capitalize]{cleveref}
\crefname{section}{Sec.}{Secs.}
\Crefname{section}{Section}{Sections}
\Crefname{table}{Table}{Tables}
\crefname{table}{Tab.}{Tabs.}

\def\cvprPaperID{11631} %
\def\confName{CVPR}
\def\confYear{2022}

\begin{document}
\newcommand{\shortname}{OnePose\xspace}

\title{\shortname: One-Shot Object Pose Estimation without CAD Models}

\author{
    Jiaming Sun$^{1,2*}$ 
    \quad Zihao Wang$^{1*}$ 
    \quad Siyu Zhang$^{2*}$ 
    \quad Xingyi He$^{1}$ 
    \quad Hongcheng Zhao$^{3}$ 
    \\
    \quad Guofeng Zhang$^{1}$ 
    \quad Xiaowei Zhou$^{1\dagger}$
    \vspace{1em}
    \\
    $^1$Zhejiang University \quad 
    $^2$SenseTime Research \quad
    $^3$TUM \quad
}

\maketitle
\blfootnote{$^*$Equal contribution. The authors from Zhejiang University are affiliated with the State Key Lab of CAD\&CG and the ZJU-SenseTime Joint Lab of 3D Vision. $^\dagger$Corresponding author: Xiaowei Zhou.}
\begin{abstract}
	We propose a new method named \shortname for object pose estimation.
	Unlike existing instance-level or category-level methods, \shortname does not rely on CAD models and can handle objects in arbitrary categories without instance- or category-specific network training.
	\shortname draws the idea from visual localization and only requires a simple RGB video scan of the object to build a sparse SfM model of the object. 
	Then, this model is registered to new query images with a generic feature matching network.
	To mitigate the slow runtime of existing visual localization methods, we propose a new graph attention network that directly matches 2D interest points in the query image with the 3D points in the SfM model, resulting in efficient and robust pose estimation.
	Combined with a feature-based pose tracker, \shortname is able to stably detect and track 6D poses of everyday household objects in real-time.
	We also collected a large-scale dataset that consists of 450 sequences of 150 objects.
	Code and data are available at the project page: \url{https://zju3dv.github.io/onepose/}.
\end{abstract}

\section{Introduction}
\label{sec:intro}
Object pose estimation plays an important role in augmented reality (AR).
The ultimate goal of object pose estimation in AR is to use arbitrary objects as ``virtual anchors'' of AR effects, 
which demands the ability to estimate poses of surrounding objects in our daily life.
Most established works in object pose estimation~\cite{xiangPoseCNNConvolutionalNeural2017, pengPVNetPixelwiseVoting2018, labbeCosyPoseConsistentMultiview2020} 
assume that the CAD model of the object is known \textit{a priori}.
Since high-quality CAD models of everyday objects are often inaccessible, the research on object pose estimation for AR scenarios necessitates new problem settings.

To not rely on instance-level CAD models, 
many recent methods have been working on category-level pose estimation~\cite{wangNormalizedObjectCoordinate2019, ahmadyanObjectronLargeScale2020}.
By training a network on different instances in the same category, 
the network can learn a category-level representation of object appearances and shapes and thus be able to generalize to new instances in the same category.
However, such approaches require a large number of training samples in the same category, which can be hard to obtain and annotate.
Furthermore, 
the generalization capabilities of category-level methods are not guaranteed when a new instance has a significantly different appearance or shape.
More importantly, training and deploying a network for each category are unaffordable in many real world applications, e.g., mobile AR, when the number of object categories to be handled is huge.

\begin{figure}[tb]
    \centering
    \includegraphics[width=\linewidth]{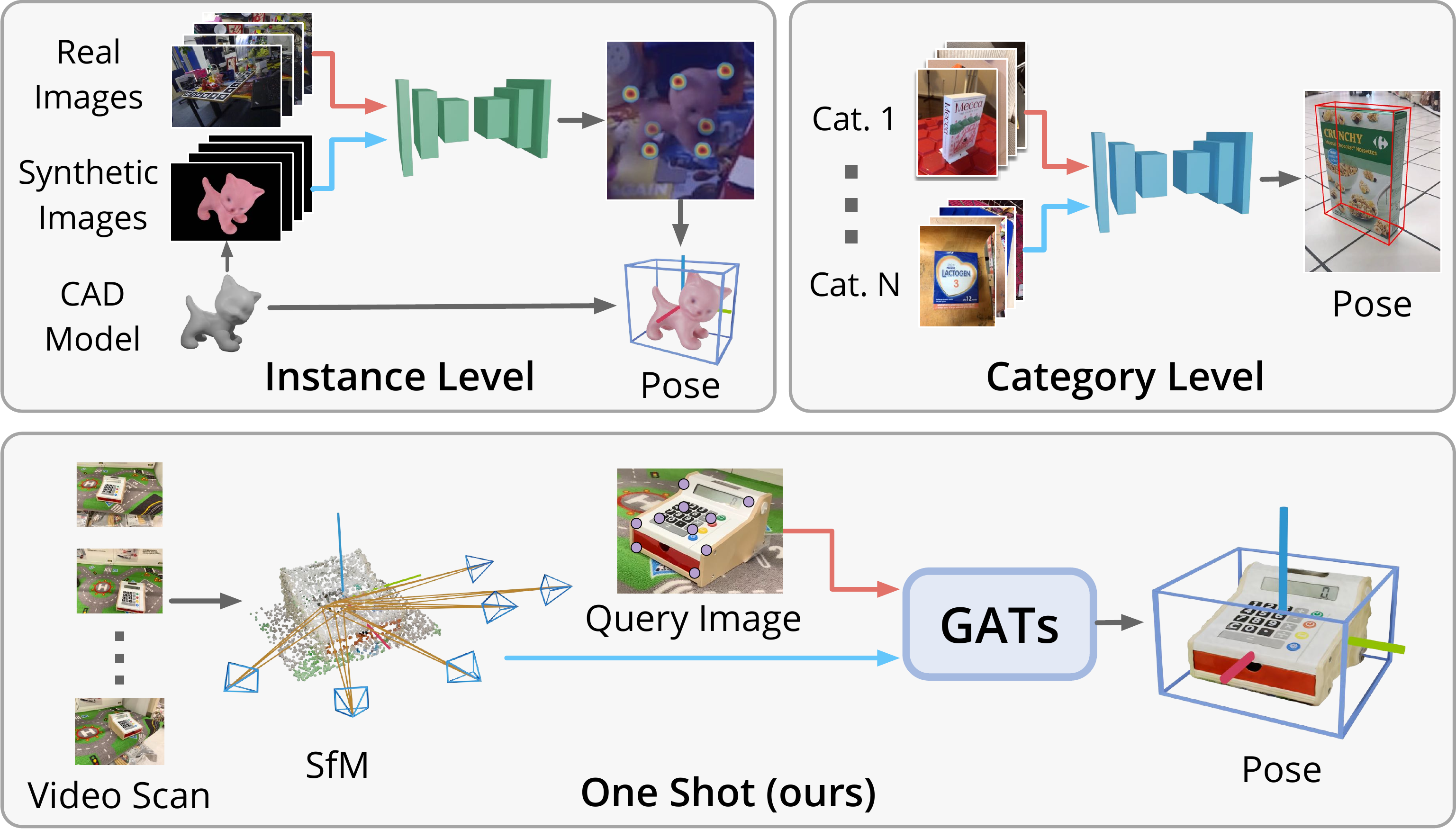}
    \caption{
    \textbf{Comparison of different problem settings} of instance/category-level object pose estimation and the one-shot pose estimation proposed in this work. Unlike previous works that rely on instance- or category-specific network training, the proposed approach only requires a simple video scan of the object to build a sparse SfM model of the object and uses a generic 3D-2D feature matching network (GATs) to estimate its pose, without CAD models or additional network training.
    }
      \vspace{-0.55 cm}
    \label{fig:figure1}
\end{figure}

To alleviate the demand for CAD models or category-specific training, we go back to an ``old'' problem setting for object pose estimation, but renovate the entire pipeline with a new learning-based approach.
Similar to the task of visual localization, which estimates the unknown camera pose given an SfM map of a scene, 
object pose estimation has long been formulated in the localization-based setting~\cite{skrypnykSceneModellingRecognition2004, martinezMOPEDScalableLow2010}.
Different from instance- or category-level methods, this setting assumes that a video sequence of the object is given, and a sparse point cloud model can be reconstructed from the sequence.
Estimating the object pose is then equivalent to localizing the camera pose with respect to the reconstructed point cloud model.
At test time, 2D local features are extracted from the query image and matched with the points in the SfM model to obtain 2D-3D correspondences, from which the object pose can be solved by PnP.
Instead of learning instance- or category-specific representations by neural networks,
this traditional pipeline leverages an explicit 3D model of the object that can be built on-the-fly for a new instance, which brings better generalization capabilities to arbitrary objects while making the system more explainable.

In this paper, we refer to this problem setting as \emph{one-shot object pose estimation}, where the objective is being able to estimate 6D pose of an object in arbitrary category, given only a few pose-annotated images of the object for training. 
While this problem is similar to visual localization,
directly migrating existing visual localization methods does not solve this problem.
The modern visual localization pipeline~\cite{sarlinCoarseFineRobust2019} produces 2D-3D correspondences by first performing a 2D-2D matching between the query image and the retrieved database images.
To ensure a high success rate of localization, matching to multiple image retrieval candidates is necessary, so that the 2D-2D matching can be expensive especially for learning-based matchers~\cite{sarlinSuperGlueLearningFeature2020,sunLoFTRDetectorFreeLocal2021}.
As a result, the runtime of existing visual localization methods is often seconds and cannot satisfy the requirement to track moving objects in real-time.

For the reasons above, we propose to directly perform 2D-3D matching between the query image and the SfM point clouds. 
Our key idea is to use graph attention networks (GATs)~\cite{velickovicGraphAttentionNetworks2018} to aggregate the
2D features that correspond to the same 3D SfM point (i.e., a feature track) to form a 3D feature.
The aggregated 3D features are later matched with 2D features in the query images with self- and cross-attention layers.
Together with the self- and cross-attention layers, the GATs can capture the globally-consented and context-dependent matching priors exhibited in ground-truth 2D-3D correspondences, making the matching more accurate and robust.

To evaluate the proposed method, we collected a large-scale dataset for the one-shot pose estimation setting, which contains 450 sequences of 150 objects.
Compared with previous instance-level method PVNet~\cite{peng2020pvnet} and category-level method Objectron~\cite{ahmadyanObjectronLargeScale2020}, \shortname achieves better precision without training for any object instances or categories in the validation set, while taking only 58 \textit{ms} to process one frame on GPU. 
To the best of our knowledge, when combined with a feature-based pose tracker, \shortname is the first learning-based method that can stably detect and track poses of everyday household objects in real-time (refer to the project page).

\paragraph{Contributions}
\vspace{-.4em}
\begin{itemize}[leftmargin=*]
\setlength\itemsep{-.3em}
\item Renovating the visual localization pipeline for object pose estimation that can handle novel objects without CAD models or additional network training.
\item A new architecture of graph attention networks for robust 2D-3D feature matching.
\item A large-scale object dataset for one-shot object pose estimation with pose annotations.
\end{itemize}

\section{Related works}
\label{sec:related}

\paragraph{CAD-Model-Based Object Pose Estimation}
The state-of-the-art approaches for the object 6DoF pose estimation can be broadly categorized into regression and keypoint techniques.
Given an image, the first type of methods~\cite{xiangPoseCNNConvolutionalNeural2017, kehl_ssd-6d_2017,labbeCosyPoseConsistentMultiview2020, ferrari_deepim:_2018} directly regress pose parameters with features within each Region of Interest (RoI).
In contrast, the latter type of methods first find correspondences between image pixels and 3D object coordinates
either by regression~\cite{pavlakos6DoFObjectPose2017, park_pix2pose:_2019, oberweger_making_2018-1} or by voting~\cite{pengPVNetPixelwiseVoting2018, peng2020pvnet}, and then compute the pose with Perspective-n-Points (PnP).
These methods require high ﬁdelity textured 3D models to generate auxiliary synthetic training data 
and for pose refinements~\cite{ferrari_deepim:_2018, labbeCosyPoseConsistentMultiview2020} to achieve high accuracy on trained instances.

Unlike the abovementioned methods that train a single network for each instance, NOCS~\cite{wangNormalizedObjectCoordinate2019} 
proposes to establish correspondences between pixels on the image and Normalized Object Coordinates (NOCS) shared within each category.
With this learned category-level shape prior, NOCS can eliminate the dependencies on CAD models during test time.
Some later works~\cite{tian_shape_2020, lee_category-level_2021, wang_category-level_2021, wang_gdr-net_2021, tian_shape_2020} follow the trend of leveraging category-level prior to further recover a more accurate shape of the object with NOCS representation.
A limitation of this line of work is that 
the shape and the appearance of some instances could vary significantly even they belong to the same category, thus the generalization capabilities of trained networks over these instances are questionable.
Moreover, accurate CAD models are still required for ground-truth NOCS map generation during training, and different networks need to be trained for different categories.
Our proposed method does not require CAD models both for training and test time and is category-agnostic.

\paragraph{CAD-Model-Free Object Pose Estimation}
Recently, a few attempts have been made to achieve CAD-model-free object 6D pose estimation both at the training and test time.
Both Neural Object Fitting~\cite{chen_category_nodate} and LatentFusion~\cite{parkLatentFusionEndtoEndDifferentiable2019} tackled the problem via analysis-by-synthesis approaches 
where differentiablly synthesized images are compared with target images to generate gradients for the object pose optimization.
Neural Object Fitting~\cite{chen_category_nodate} proposes to encode category-level appereance prior with a Variational Auto Encoder (VAE) trained with fully synthetic data,
while LatentFusion~\cite{parkLatentFusionEndtoEndDifferentiable2019} builds a 3D latent space based object representation with posed RGB-D images for each unseen object.
However, the efficiency and accuracy of such methods are highly limited by image synthesizing networks and are not suitable for AR applications. 
RLLG~\cite{caiReconstructLocallyLocalize} takes a different approach and learns correspondences from image pixels to object coordinates~\cite{fleet_learning_2014} without CAD models.
Although RLLG can achieve comparable precision to its counterparts~\cite{park_pix2pose:_2019}, 
it works only on the instance level and requires highly accurate instance masks to segment foreground pixels.
Most recently, Objectron~\cite{ahmadyanObjectronLargeScale2020} proposes a data-driven approach that learns to regress pixel coordinates of projected box corners for each category with a tremendous amount of annotated training data.
Such an approach is costly and only limited to a few categories as the learned model is category-specific. 
Moreover, it can only obtain up-to-scale poses without metric scales since it uses a single-view image as input.
On the contrary, our method can leverage the visual-inertial odometry to recover metric scales during the mapping stage, thus being able to recover metric 6D poses at test time.

\paragraph{Feature-Matching-Based Pose Estimation}
Visual localization pipelines based on feature-matching have long been studied.
Traditionally, the localization problem is solved by finding 2D-3D correspondences between input RGB images and a 3D model from SfM with hand-crafted local features like  
SIFT~\cite{loweDistinctiveImageFeatures2004}, FAST~\cite{leonardis_machine_2006} and ORB~\cite{rubleeORBEfficientAlternative2011}. 
Recently, learning-based local feature detection, description~\cite{detoneGeometricDeepSLAM2017,detoneSuperPointSelfSupervisedInterest2018,dusmanuD2NetTrainableCNN2019,tyszkiewiczDISKLearningLocal2020} and matching~\cite{sarlinSuperGlueLearningFeature2020,sunLoFTRDetectorFreeLocal2021} surpass these hand-crafted methods and have substituted the traditional counterparts in the localization pipeline.
Notably, Hierarchical Localization (HLoc)~\cite{sarlinCoarseFineRobust2019} provides a complete toolbox for running SfM with COLMAP~\cite{schonbergerStructurefromMotionRevisited2016} and feature extraction and matching with SuperGlue~\cite{sarlinSuperGlueLearningFeature2020}.
Our method is inspired by SuperGlue in terms of using self- and cross-attention layers for feature matching. 
However, SuperGlue only focuses on 2D-2D matching between images and does not consider the graph structure of the SfM map.
Our method uses graph attention networks~\cite{velickovicGraphAttentionNetworks2018} to process and aggregate 2D features that correspond to a 3D SfM point (i.e., a feature track), which preserves the graph structure of the SfM during 2D-3D matching.

Many traditional methods for object recognition and pose estimation also share the feature-based pipeline similar to visual localization.
These methods first build object models by reconstructing sparse point clouds from matched keypoints across the views~\cite{skrypnykSceneModellingRecognition2004,Collet2009ObjectRA,martinezMOPEDScalableLow2010,tangTexturedObjectRecognition2012}, and localize with the sparse point cloud model given a query image.
Some approaches~\cite{qiuTracking3DMotion2019,wenBundleTrack6DPose2021} propose to build a point cloud model online with a framework similar to Simultaneous Localization and Mapping (SLAM). 
Notably, BundleTrack~\cite{wenBundleTrack6DPose2021} proposes an online pose tracking pipeline without instance- or category-level models, which resembles ours mostly.
However, it uses 2D-2D feature matching instead of 2D-3D as in ours.  To recover the 3D information, it also takes depth map as input which could limit its usage in AR.

\begin{figure*}
\vspace{-0.8 cm}
\begin{center}
\includegraphics[width=.8\linewidth]{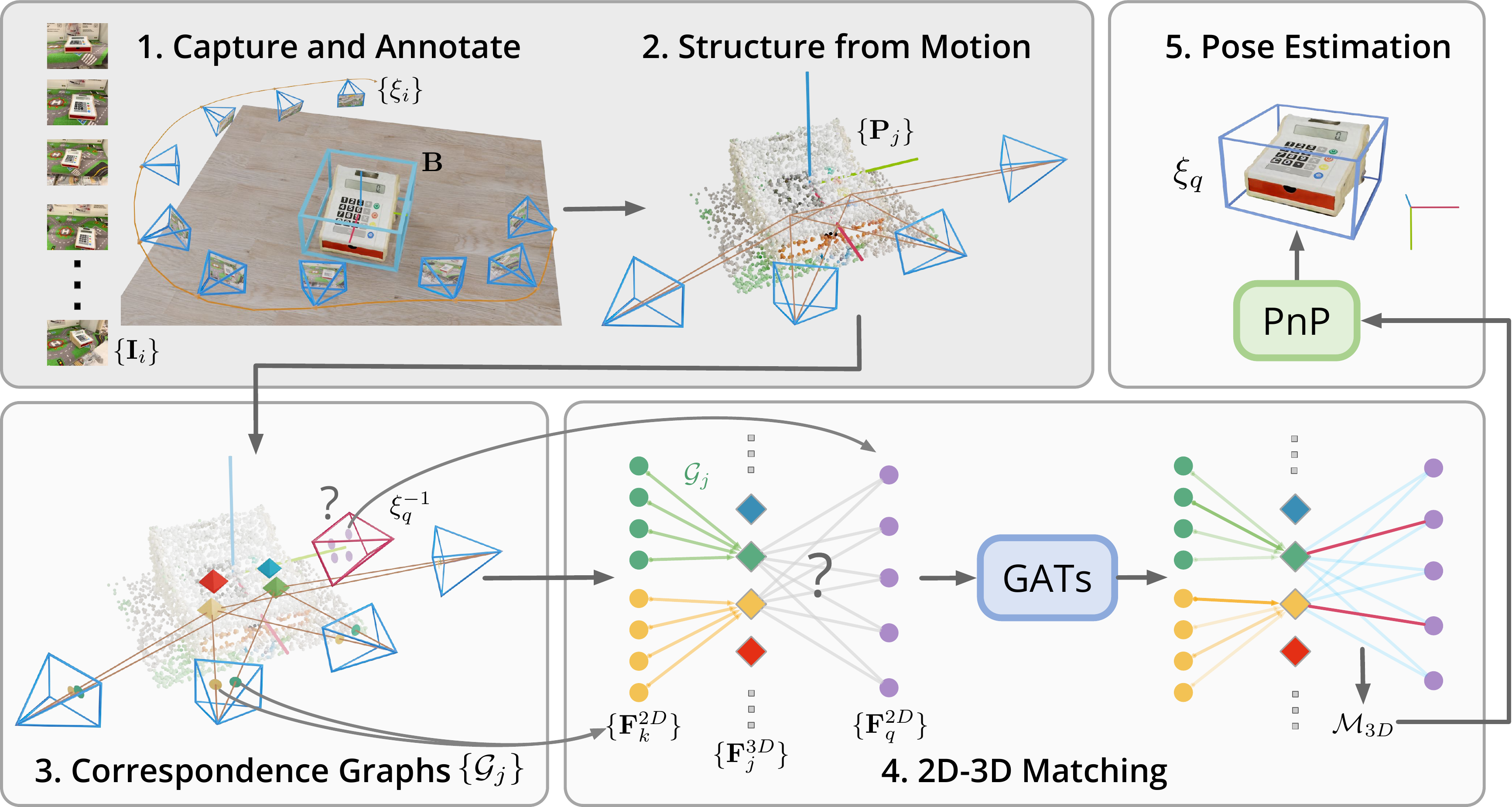}
\end{center}
\vspace{-0.2 cm}
\caption{
\textbf{Overview of \shortname.}
\textbf{1.} For each object, a video scan with RGB frames $\{\mathbf{I}_i\}$ and camera poses $\{\xi_{i}\}$ are collected together with the annotated 3D object bounding box $\mathbf{B}$.
\textbf{2.} Structure from Motion (SfM) reconstructs a sparse point cloud $\{\mathbf{P}_j\}$ of the object.
\textbf{3.} The correspondence graphs $\{\mathcal{G}_j\}$ (\protect\scalerel*{\includegraphics{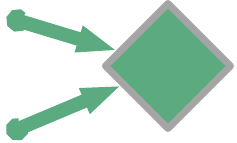}}{B},\protect\scalerel*{\includegraphics{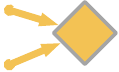}}{B}) are built during SfM, which represent the 2D-3D correspondences in the SfM map. \protect\scalerel*{\includegraphics{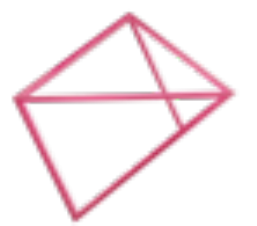}}{B} represents the camera to be localized in the object frame.
\textbf{4.} 2D descriptors $\{\mathbf{F}_k^{2D}\}$ (\protect\scalerel*{\includegraphics{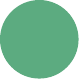}}{B},\protect\scalerel*{\includegraphics{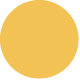}}{B}) are aggregated to 3D descriptors $\{\mathbf{F}_j^{3D}\}$ (\protect\scalerel*{\includegraphics{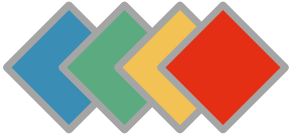}}{B}) with the aggregration-attention layer. $\{\mathbf{F}_j^{3D}\}$ are later matched with 2D descriptors from the query image $\{\mathbf{F}_q^{2D}\}$ (\protect\scalerel*{\includegraphics{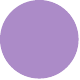}}{B}) to generate 2D-3D match predictions $\mathcal{M}_{3D}$.
\textbf{5.} Finally, the object pose $\xi_{q}$  is computed by solving the PnP problem with $\mathcal{M}_{3D}$.
Grey background color denotes offline processes. Best viewed in color.
}
  \vspace{-0.2 cm}
\label{fig:overview}
\end{figure*} %

\section{Method}
\label{sec:method}
An overview of the proposed method is presented in Fig.~\ref{fig:overview}.
In the setting of one-shot object pose estimation introduced in Sec.~\ref{sec:intro}, a video scan surrounding the object is captured with a mobile device (e.g. iPhone or iPad).
Given the video scan and a test image sequence $\{\mathbf{I}_q\}$, the objective of one-shot object pose estimation is to estimate the object poses $\{\xi_q\} \in \mathbb{SE}(3)$ defined in the camera coordinate system, where $q$ is the key-frame index in the video.
We use bold letters (e.g. $\mathbf{I}$) to denote tensors, calligraphy letters (e.g. $\mathcal{G}$) to denote graphs and $\{\cdot\}$ to denote a set of these entities.

\subsection{Preliminaries}
\begin{figure}[ht]
    \vspace{-0.8 cm}
    \includegraphics[width=0.9\linewidth]{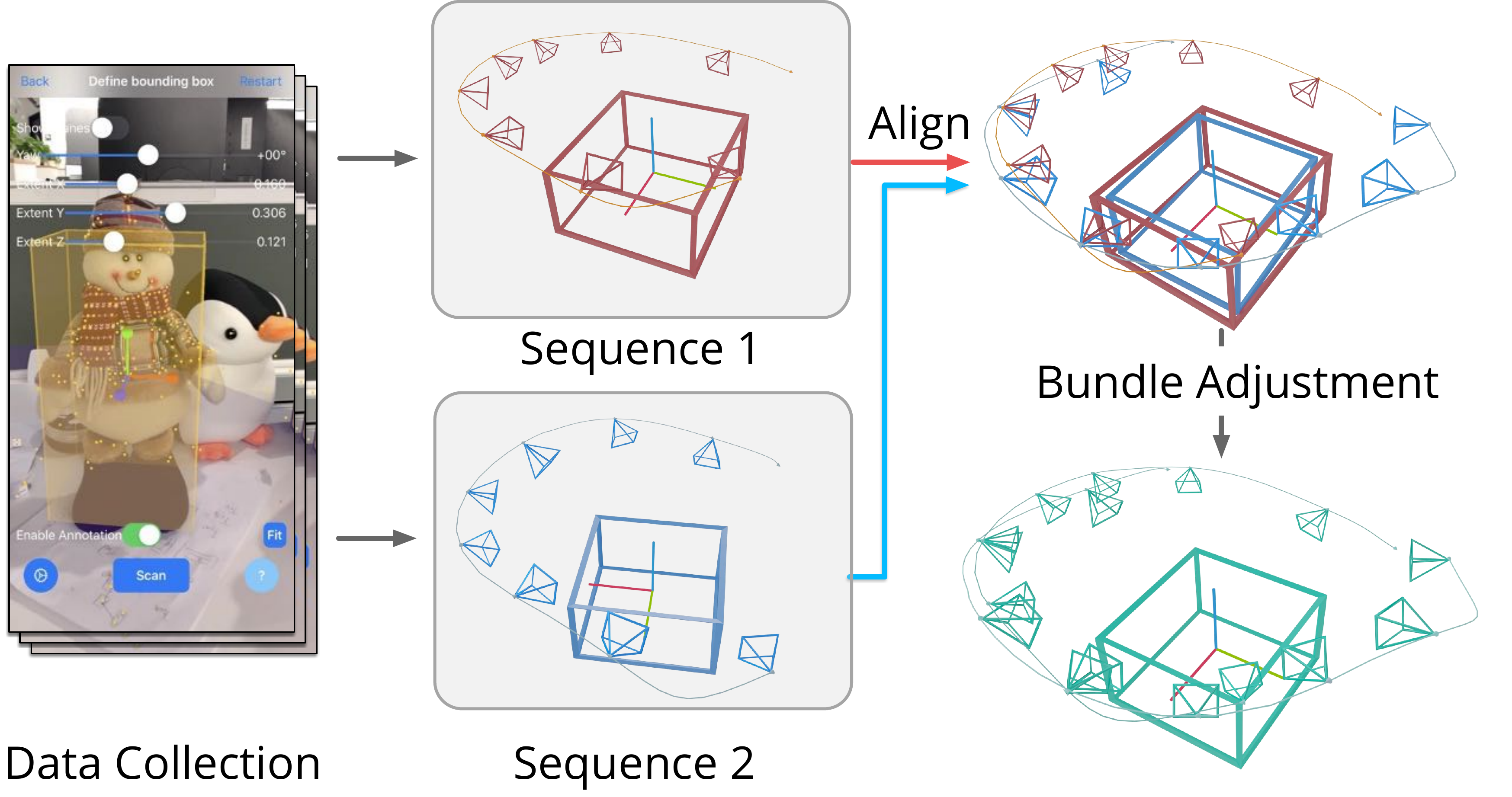}
    \caption{
        \textbf{Dataset collection and sequence registration by joint bundle adjustment.} 
        The AR-based dataset collection and annotation interface are shown on the left.
        Multiple collected sequences of the same object are aligned according to the bounding box annotations. 
        The final camera poses are optimized by bundle adjustment with the camera poses from ARKit as prior.
    }
    \label{fig:ba_match}
\end{figure}

\paragraph{Data Capture and Annotation}
During the data capture, the object is assumed to be set on a flat surface and remains static during the capture. 
To define the canonical pose of the object, an object bounding box $\mathbf{B}$ is annotated in AR, 
with the camera poses $\{\xi_{i}\} \in \mathbb{SE}(3)$ tracked by off-the-self AR toolboxes like ARKit~\cite{arkit} or ARCore~\cite{arcore}. 
$i$ is the frame index.
The capture interface is shown in Fig.~\ref{fig:ba_match}.
$\mathbf{B}$ is parameterized by the center location, dimensions and rotation around the \emph{z}-axis (yaw angle).
After the data capture and annotation, the pipeline of \shortname can be separated into the offline mapping phase and the online localization phase.

\paragraph{Structure from Motion}
In the mapping phase, given a set of images $\{\mathbf{I}_i\}$ extracted from the video scan, 
we use Structure from Motion (SfM) to reconstruct the sparse point cloud $\{\mathbf{P}_j\}$ of the object, where $j$ is the point index. 
Since $\mathbf{B}$ is annotated, $\{\mathbf{P}_j\}$ can be defined in the object coordinate system.
A visualization of all correspondence graphs of the object $\{\mathcal{G}_j\}$ can be found in Fig.~\ref{fig:overview} (\textbf{3},\textbf{4}).
Specifically, 2D keypoints and descriptors are first extracted from each image and matched between images to produce 2D-2D correspondences.
Every reconstructed point $\mathbf{P}_j$ corresponds to a set of matched 2D keypoints and descriptors $\{\mathbf{F}_k^{2D}\}\in\mathbb{R}^{d}$ where $k$ is the keypoint index and $d$ is the dimension of the descriptor.
The correspondence graphs $\{\mathcal{G}_j\}$, which are also called the feature tracks, are formed by keypoint indexes for $\{\mathbf{P}_{j}^{3D}\}$ as visualized in Fig.~\ref{fig:overview} (\textbf{3},\textbf{4}).

\paragraph{Pose Estimation through Visual Localization}
In the localization phase, a sequence of query images $\{\mathbf{I}_q\}$ are captured in real-time.
Localizing the camera poses of the query images $\{\xi_{q}^{-1}\}$ with respect to $\{\mathbf{P}_j\}$ produces the object poses $\{\xi_{q}\}$ defined in the camera coordinate.

For each $\mathbf{I}_q$, 2D keypoints and descriptors $\{\mathbf{F}_q^{2D}\}\in\mathbb{R}^{d}$ are extracted and used for matching.
In modern visual localization pipeline~\cite{sarlinCoarseFineRobust2019}, an image retrieval network is used to extract image-level global features, which can be used to retrieve the image candidates from the SfM database for 2D-2D matching.
Increasing the number of image pairs to be matched will significantly slow down the localization, especially for learning-based matchers like SuperGlue~\cite{sarlinSuperGlueLearningFeature2020} or LoFTR~\cite{sunLoFTRDetectorFreeLocal2021}.
Reducing the number of images retrieved can result in a low localization success rate and thus a trade-off has to be made between runtime and pose estimation accuracy.

To remedy this problem, we propose to directly perform 2D-3D matching between the query image and the SfM point clouds.
Direct 2D-3D matching avoids the need of the image retrieval module, and thus can maintain localization accuracy while being fast.
In the next section, we describe how to obtain the 2D-3D correspondences $\mathcal{M}_{3D}$.

\subsection{\shortname}
\paragraph{Graph Attention Networks (GATs) for 2D-3D Matching}
Direct 2D-3D matching requires 3D feature descriptors.
Since a 3D point $\mathbf{P}_j$ is associated with multiple $\mathbf{F}_{k\prime}^{2D}$ in $\mathcal{G}_j$, an aggregation operation is needed to update the 3D descriptors, defined as $\{\mathbf{F}_j^{3D}\}\in\mathbb{R}^{d}$ which are initialized by averaging the corresponding 2D descriptors.
The aggregation operation could cause information loss since it reduces multiple descriptors to one.
An ideal aggregation operation should be able to adaptively preserve the most informative 2D features in $\{\mathbf{F}_{k}^{2D}\}$ for the 2D-3D matching according to different $\mathbf{F}_q^{2D}$.

We propose to use the graph attention layer in~\cite{velickovicGraphAttentionNetworks2018} to achieve the adaptive aggregation.
We name it the aggregation-attention layer.
The aggregation-attention layer operates on each individual $\mathcal{G}_j$. 
For every $\mathcal{G}_j$, denoting the weight matrix as ${\mathbf W} \in \mathbb{R}^{d \times d}$, the aggregation-attention layer is defined as:
\begin{equation*}
\resizebox{0.8\hsize}{!}{
	$
	\operatorname{Aggr}(\{\mathbf{F}_{k^\prime}^{2D}\}, \mathbf{F}_j^{3D}) = \mathbf{F}_j^{3D} + \underset{\forall k^\prime \in \mathcal{G}_j}{\sum} \alpha_{k^\prime} \mathbf{{F}}_{k^\prime}^{2D},
	$
}
\end{equation*}
\vspace{-1em}
\begin{equation*}
\resizebox{0.74\hsize}{!}{
	$
\alpha_{k^\prime} = \underset{\forall k^\prime\in \mathcal{G}_j}{\operatorname{softmax}}(\operatorname{sim} (\mathbf{W} \cdot \mathbf{F}_{k^\prime}^{2D}, \mathbf{W} \cdot \mathbf{F}_j^{3D}))
	$
}
\end{equation*}
 with $\operatorname{sim}(\cdot, \cdot) = \langle\mathbb{R}^{d} , \mathbb{R}^{d} \rangle \in \mathbb{R}$ computes the attention coefficient, which measures the importance of the descriptors in the aggregation operation.%

Inspired by \cite{sarlinSuperGlueLearningFeature2020,sunLoFTRDetectorFreeLocal2021}, 
we further use self- and cross-attention layers following the aggregation-attention layers to process and transform the aggregated 3D descriptors and query 2D descriptors.
A set of aggregation-, self- and cross-attention layers forms an \emph{attention group}, specifically:
\begin{equation*}
\resizebox{0.8\hsize}{!}{
	$
	\begin{cases}
		\{\mathbf{\hat{F}}_j^{3D}\} = \operatorname{Aggr}(\{\mathbf{F}_{k}^{2D}\}, \{\mathbf{F}_j^{3D}\}),\\
		\{\mathbf{\tilde{F}}_q^{2D}\} = \operatorname{Self}(\{\mathbf{F}_q^{2D}\}, \{\mathbf{F}_q^{2D}\}),\\ 
		\{\mathbf{\tilde{F}}_j^{3D}\} = \operatorname{Self}(\{\mathbf{\hat{F}}_j^{3D}\}, \{\mathbf{\hat{F}}_j^{3D}\}), \\
		\{\mathbf{F^\prime}_q^{2D}\}, \{\mathbf{F^\prime}_j^{3D}\} = \operatorname{Cross}(\{\mathbf{\tilde{F}}_q^{2D}\}, \{\mathbf{\tilde{F}}_j^{3D}\}).
	\end{cases}
	$
}
\end{equation*}

\begin{figure}[ht]
    \vspace{-0.8 cm}
    \includegraphics[width=\linewidth]{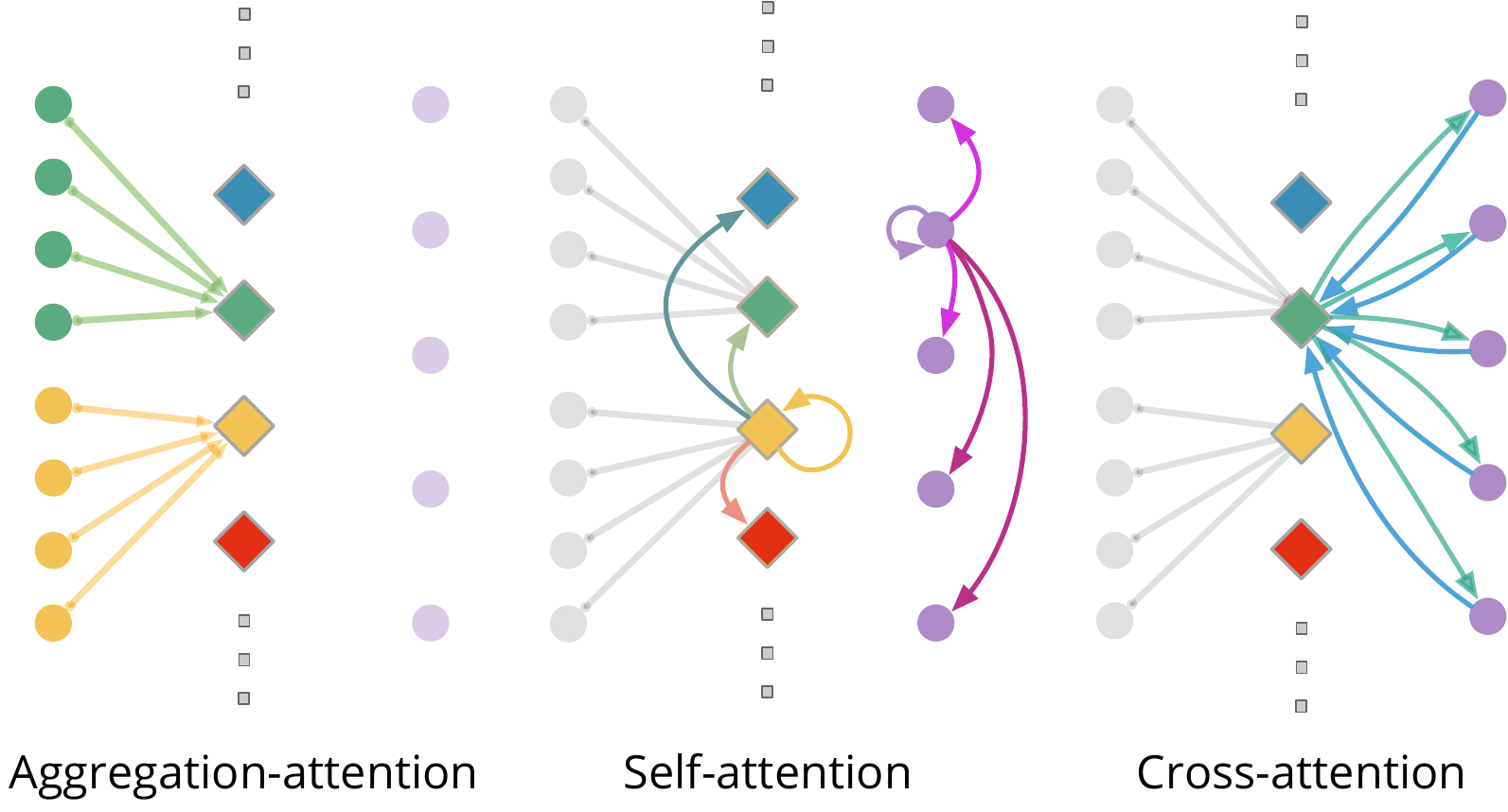}
    \caption{
        \textbf{Different types of attention layers in GATs.} 
        $\{\mathbf{F}_j^{3D}\}$ :        \protect\scalerel*{\includegraphics{fig/orig_images/gats_caption1.pdf}}{B},
        $\{\mathbf{F}_q^{2D}\}$ :        \protect\scalerel*{\includegraphics{fig/orig_images/gats_caption2.pdf}}{B},
        $\{\mathcal{G}_j\}$     :    (\protect\scalerel*{\includegraphics{fig/orig_images/gats_caption3_green.pdf}}{B},\protect\scalerel*{\includegraphics{fig/orig_images/gats_caption3_yellow.pdf}}{B})
        $\{\mathbf{F}_k^{2D}\}$: (\protect\scalerel*{\includegraphics{fig/orig_images/caption2_green.pdf}}{B},\protect\scalerel*{\includegraphics{fig/orig_images/caption2_yellow.pdf}}{B}).
        For clarity, the complete relations of attentions in the figure are not shown.
    }
    \label{fig:gats_layer}
\end{figure}

The proposed architecture of graph attention networks (GATs) is composed of $N$ stacked attention groups. 
Intuitively, the aggregation-attention layers will adaptively attend to different $\mathbf{F}_{k}^{2D}$ in $\mathcal{G}_j$ according to its relevance with $\mathbf{F}_q^{2D}$, thus preserving more descriminative information for 2D-3D matching.
By interleaving the aggregation-attention layers with self- and cross-attention layers, $\{\mathbf{F}_k^{2D}\}$, $\{\mathbf{F}_j^{3D}\}$, $\{\mathbf{F}_q^{2D}\}$ can exchange information with each other, thus making the matching globally-consented and context-dependent.

\paragraph{Match Selection and Pose Calculation}
We follow~\cite{sunLoFTRDetectorFreeLocal2021} to use the dual-softmax operator to differentiablly extract match confidence scores $\mathcal{P}_{3D}$.
The score matrix $\mathbf{S}$ between the transformed features is first calculated by $\mathbf{S}\left(q, j\right) = \langle\mathbf{F^\prime}_q^{2D}, \mathbf{F^\prime}_j^{3D}\rangle$.
Formally, the matching confidence $\mathbf{C}_{3D}$ is obtained by:
\begin{equation*}
    \mathbf{C}_{3D}(q, j) = \operatorname{softmax}\left(\mathbf{S}\left(q, \cdot \right)\right)_j \cdot \operatorname{softmax}\left(\mathbf{S}\left(\cdot, j\right)\right)_q.
    \label{eq:dual-softmax}
\end{equation*}
After selecting a confidence threshold $\theta$, $\mathbf{C}_{3D}$ becomes a permutation matrix $\mathcal{M}_{3D}$, which represents the 2D-3D match predictions.
With $\mathcal{M}_{3D}$, the object pose in the camera coordinate $\xi_{q}$ can be computed by the Perspective-n-Point (PnP) algorithm with RANSAC.

\paragraph{Supervision}
The supervision signal $\mathcal{M}_{3D}^{gt}$ can be directly obtained from filtered 2D-3D correspondences in the SfM maps in the training set.
The loss function $L$ is the focal loss~\cite{linFocalLossDense2018} over the confidence scores $\mathbf{C}_{3D}$ returned by the dual-softmax operator:
\begin{equation*}
 	L = - (1 - \mathbf{C}^\prime_{3D}\left(q, j\right)) ^ \gamma \log \mathbf{C}^\prime_{3D}\left(q, j\right),
	\label{eq:loss_coarse}
\end{equation*}
\vspace{-1.4em}
\begin{equation*}
\resizebox{.8\hsize}{!}{
	$
	\begin{cases}
		\mathbf{C}^\prime_{3D}(q, j) = \mathbf{C}_{3D}(q, j) &\text{if} \: \mathcal{M}_{3D}^{gt}(q, j) = 1 \\
		\mathbf{C}^\prime_{3D}(q, j) = 1 - \mathbf{C}_{3D}(q, j) &\text{if} \: \mathcal{M}_{3D}^{gt}(q, j) \neq 1.
	\end{cases}
	$
}
\end{equation*}

\paragraph{Online Feature-based Pose Tracking}
The above-mentioned pose estimation module takes only sparse key-frame images as input.
To obtain stable object poses for AR applications, we further equip \shortname with a feature-based pose tracking module,
which processes \emph{every} frame in the test sequence.
Similar to a SLAM system, the pose tracking module reconstructs a 3D map online and maintains its own key-frame pool.
At each time-step, tracking adopts a tightly-coupled approach and relies on both the prebuilt SfM map and the online-built 3D map to find 2D-3D correspondences and solve for 6D poses.
Since the pose tracking module preserves 2D and 3D information of the test sequence in the online-built map, it can be more stable than the single-frame-based pose estimation module.
The pose estimation module helps to recover and re-initialize the tracking module when it fails.
We provide more details about the pose tracking module in the supplementary material.

\paragraph{Remarks on the One-Shot Setting}
Other than not using CAD models or additional network training, the one-shot setting of \shortname has many advantages compared with existing instance- or category-level pose estimation methods.
During the mapping phase, \shortname takes as input a simple video scan of an object and builds an instance-specific 3D representation of the object geometry.
Similar to the role of CAD models in instance-level methods, the 3D geometry of the object is crucial for recovering object poses with metric scales.
In the localization phase, 
learned local feature matching in \shortname can handle large changes in viewpoint, lighting and scale, making the system more stable and robust compared to category-level methods.
The local-feature-based pipeline also allows the pose estimation module to be naturally coupled with a feature-based tracking module to realize efficient and stable pose tracking.

\subsection{\shortname Dataset}
Since there is no existing large-scale dataset that can fit the setting of one-shot pose estimation, 
we collected a dataset with multiple video scans of the same object put in different locations.
The \shortname dataset contains over 450 video sequences of 150 objects.
For each object, multiple video recordings, accompanied camera poses and 3D bounding box annotations are provided.
These sequences are collected under different background environments, and each has an average recording-length of 30 seconds covering all views of the object.
The dataset is randomly divided into training and validation sets.
For each object in the validation set, we assign one mapping sequence for building the SfM map, and use a test sequence for the evaluation. 

To reduce the manual labor of data annotation, we propose a semi-automatic approach to simultaneously collect and annotate the data in AR. 
To be specific, an adjustable 3D bounding box is rendered onto the image in AR, as shown in Fig.~\ref{fig:ba_match}.
The only manual work is to adjust the rotation and rough dimensions of the 3D bounding box.
Visualizations of the data capture interface and the post-processing process are shown in Fig.~\ref{fig:ba_match}.

The objective of the post-processing is to reduce the pose drift error of ARKit for each sequence and ensure consistent pose annotations across sequences. To achieve this, we first align sequences with the annotated bounding boxes and perform bundle adjustment (BA) with COLMAP~\cite{schonbergerStructurefromMotionRevisited2016,sarlinCoarseFineRobust2019}.  
Feature matches used in the BA are extracted with SuperGlue.
As the backgrounds are different between sequences, 
we extract matches only in the foreground (i.e., within the 2D object bounding boxes) between all matchable pairs of images. 
For more details about our data collection and processing pipeline, please refer to our supplementary material.

\begin{table*}[ht]
\vspace{-0.7cm}
    \centering
    \resizebox{0.95\textwidth}{!}{

\begin{tabular}{c||ccc|ccc|ccc|c} 
\Xhline{3\arrayrulewidth}

\multirow{2}{*}{\begin{tabular}[c]{@{}c@{}}\\\end{tabular}} & \multicolumn{3}{c|}{Large Objects}                        & \multicolumn{3}{c|}{Medium Objects}              & \multicolumn{3}{c|}{Small Objects}               & \multicolumn{1}{l}{\multirow{2}{*}{Time (\textit{ms})}}  \\ 
\cline{2-10}
                                                            & 1cm-1deg       & 3cm-3deg       & 5cm-5deg                & 1cm-1deg       & 3cm-3deg       & 5cm-5deg       & 1cm-1deg       & 3cm-3deg       & 5cm-5deg       & \multicolumn{1}{l}{}                                     \\ 
\hline
HLoc~\textit{(SIFT + NN)}                                   & 0.314          & 0.572          & 0.572                   & 0.432          & 0.575          & 0.608          & 0.248          & 0.468          & 0.515          & 116.27                                                   \\
HLoc~\textit{(SPP + NN)}                                    & 0.357          & 0.675          & 0.675                   & 0.508          & 0.659          & 0.706          & 0.342          & 0.612          & 0.687          & 136.98                                                   \\
HLoc~\textit{(SPP + SPG)}                                   & 0.435          & 0.813          & 0.813                   & \textbf{0.643} & 0.793          & 0.831          & \textbf{0.432} & \textbf{0.739} & \textbf{0.837} & 618.29                                                   \\
Ours                                                        & \textbf{0.471} & \textbf{0.856} & \textbf{\textbf{0.856}} & 0.629          & \textbf{0.816} & \textbf{0.858} & 0.405          & 0.729          & 0.832          & \textbf{58.31}                                           \\
\Xhline{3\arrayrulewidth}
\end{tabular}
}
\caption{\textbf{Comparison with the \textit{Visual Localization} baselines.} 
Our method are compared with HLoc\cite{sarlinCoarseFineRobust2019} with different detectors including \textit{SIFT}\cite{loweDistinctiveImageFeatures2004} and SuperPoint~(\textit{SPP})\cite{detoneSuperPointSelfSupervisedInterest2018}, and matchers including Nearest Neighbor (\textit{NN}) and SuperGlue (\textit{SPG})\cite{sarlinSuperGlueLearningFeature2020}.
}
\label{tab:exp_visloc}
\end{table*}
\begin{table}[htb]
\centering
\resizebox{0.7\columnwidth}{!}{%
\setlength\tabcolsep{2pt} %
\begin{tabular}{c||ccccccc} 
 \Xhline{3\arrayrulewidth}
Obj. ID & 0447           & 0450           & 0488           & 0493           & 0494           & 0524           & 0594            \\ 
\hline
PVNet  & 0.253          & 0.127          & 0.042          & 0.094          & 0.192          & 0.119          & 0.077           \\
Ours   & \textbf{0.900} & \textbf{0.981} & \textbf{0.740} & \textbf{0.873} & \textbf{0.819} & \textbf{0.679} & \textbf{0.789}  \\
 \Xhline{3\arrayrulewidth}
\end{tabular}
}
\vspace{-0.15cm}
\caption{\textbf{Comparison with the \textit{instance-level} baseline.}
Our method is compared with PVNet\cite{pengPVNetPixelwiseVoting2018}
 on selected objects from the \shortname dataset with the \textit{5cm-5deg} metric.} 
  \vspace{-0.3 cm}
\label{tab:exp_pvnet}
\end{table}

\section{Experiments}
In this section, we first introduce our selection of baseline methods and evaluation protocols,
as well as evaluation metrics on our proposed \shortname dataset in Sec.~\ref{sub_sec:exp_intro}, followed by implementation details of our method in Sec.~\ref{sub_sec:imp_details}.
Experimental results and ablation studies are detailed in Sec.~\ref{sub_sec:main_result} and Sec. \ref{sub_sec:ablation}, respectively.

\subsection{Experiment Settings and Baselines}
\label{sub_sec:exp_intro}

\paragraph{Baselines}
We compare our method with the following baseline methods in three categories:
1) \textit{Visual Localization} methods are most relevant to the proposed method in terms of estimating the pose based on local feature matching.
To be specific, we compare our method with HLoc~\cite{sarlinCoarseFineRobust2019} using different 
keypoint descriptors (SIFT~\cite{loweDistinctiveImageFeatures2004} and SuperPoint~\cite{detoneSuperPointSelfSupervisedInterest2018}), 
as well as matchers (Nearest Neighbour, SuperGlue~\cite{sarlinSuperGlueLearningFeature2020}).
2) \textit{Instance-level} method PVNet~\cite{pengPVNetPixelwiseVoting2018,peng2020pvnet}.
3) \textit{Category-level} method Objectron~\cite{ahmadyanObjectronLargeScale2020}.
To the best of our knowledge, Objectron is the only method for category-level object pose estimation with RGB image as input. 

\paragraph{Evaluation Protocols}
We apply per-frame pose estimation with the proposed method without the pose tracking module for a fair comparison in all the experiments.
For our \textit{Visual Localization} baselines and the proposed method, we use the same video scan to build 
the SfM map for the localization.
Note that the original image retrieval module used for large scale scenes does not generalize well to objects,
thus we equally sample a subset of five images with equal intervals from database images as retrieved images for feature matching.
To train our \textit{instance-level} baseline PVNet, 
we use 3D box corners instead of sampled semantic points from CAD models as keypoints to vote for, 
and further supply auxiliary mask supervision which is indispensable for training PVNet.
Due to the data demanding nature of the \textit{category-level} baseline Objectron~\cite{ahmadyanObjectronLargeScale2020}, 
we directly use the models provided by the authors, which are trained on the original Objectron dataset.

\paragraph{Metrics}
For evaluation metrics, we cannot directly adopt the commonly used ADD metric~\cite{hutchison_model_2013} and 2D 
projection metric~\cite{Brachmann_2016_CVPR} since CAD models are unavailable in our setting.
Another commonly used metric for evaluating the quality of predicted object pose is the \textit{5cm-5deg} metric proposed in \cite{shotton_scene_2013} 
which deems a predicted pose as correct if the error is below 5\textit{cm} and 5$\degree$.
We further narrow down the criteria to \textit{1cm-1deg} and \textit{3cm-3deg} following a similar definition to set up more strict metrics for the pose estimation in augmented reality application.
We divide the objects to three splits by their diameters with 40 \textit{cm} and 25 \textit{cm} as thresholds.
When comparing with instance-level baseline and category-level baseline, we follow the metrics used in the original paper.

\subsection{Implementation Details}
\label{sub_sec:imp_details}
During the mapping phase, to maintain a fast mapping speed, we reuse $\{\xi_{i}\}$ and use triangulation to reconstruct the point cloud, without further optimization on the camera poses by bundle adjustment.
During the localization phase, we assume the 2D bounding box of the object is known, which can be easily obtained from an off-the-shelf 2D object detector (e.g. YOLOv5~\cite{UltralyticsYolov52021}) in practice.
To reduce possible mismatches in pose estimation, only the 3D points inside the annotated 3D bounding box are preserved during mapping,
and only the 2D features inside the detected 2D bounding box are preserved during localization.
For the network design, we use $N=4$ attention groups in GATs.
Linear Attention~\cite{katharopoulos_et_al_2020, vyas_et_al_2020} is used in all the attention layers following ~\cite{sunLoFTRDetectorFreeLocal2021}.
As the input of GATs, we randomly sample or pad a set of eight features from $\{\mathbf{F}_i^{2D}\}$ associated with each $\mathbf{F}_i^{3D}$
for all experiments in the paper.
The  $\{\mathbf{F}_i^{3D}\}$ are initialized by averaging all of the associated features $\{\mathbf{F}_i^{2D}\}$.

\subsection{Evaluation Results}
\label{sub_sec:main_result}

\paragraph{Comparison with Visual Localization Baselines}
We compare our approach with visual localization baselines with different feature extractors and matchers, and present the results in Tab.~\ref{tab:exp_visloc}.
HLoc~\textit{(SPP + SPG)} is the baseline with learning-based feature extractor (SuperPoint) and matcher (SuperGlue), which mostly resembles our method among all the three variants.
Our method performs on-par or slightly better compared with 
HLoc~\textit{(SPP + SPG)}, 
while HLoc~\textit{(SPP + SPG)} takes ten times the runtime of our method.
We believe the improvement comes from the ability of our method to selectively aggregate context from multiple images benefited from our GATs design,
instead of only focusing on the two images being matched.

\paragraph{Comparison with the Instance-level Baseline PVNet} 
The proposed method is compared with PVNet~\cite{pengPVNetPixelwiseVoting2018} with \textit{5cm-5deg} on selected objects from our \shortname dataset 
and the results are as presented in Tab.~\ref{tab:exp_pvnet}.
To obtain segmentation masks for training PVNet, we need to additionally apply dense 3D reconstruction and render the reconstructed meshes to obtain masks on the data sequences.
This process is time-consuming and greatly limits our choices for objects because of the quality of 3D reconstruction.
Our method achieves much higher precision than PVNet, which demonstrates the superiority of our method.
PVNet relies on memorizing the mapping from image patches to object-specific keypoints.
Without pre-training on large-scale synthetic images (rendered with CAD models) that densely cover all possible views, the performance of PVNet drops drastically. 
Conversely, our method is able to leverage the learned local features that are relatively viewpoint-invariant and thus generalize to unseen views while maintaining the precision.

\begin{figure*}[btp]
	\vspace{-1.7em}
	\centering
	\includegraphics[width=0.85\linewidth]{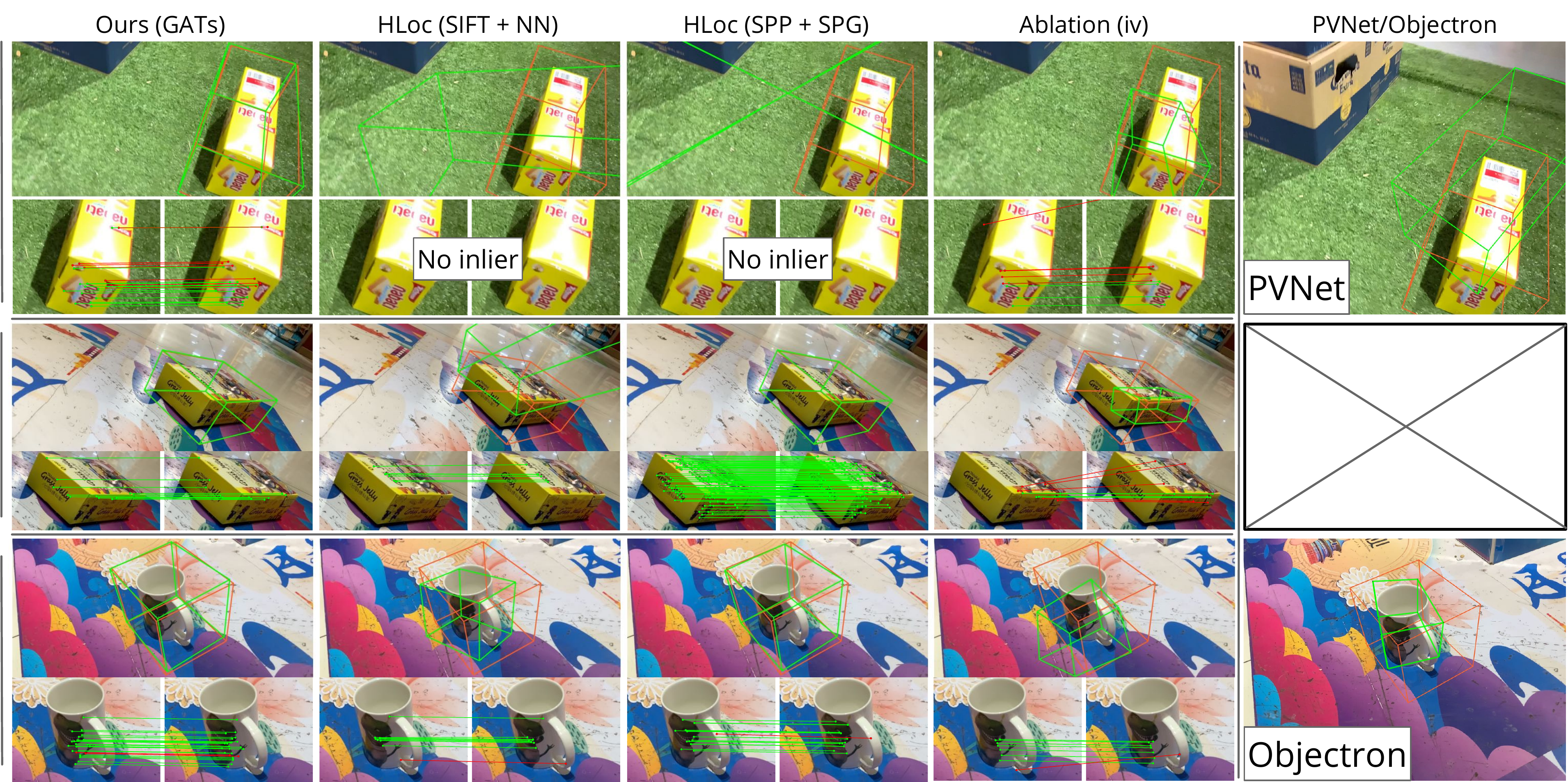}%
	\caption{{\textbf{Qualitative results.}}
	Every two rows show the results on one test image.
	Green bounding boxes denote predictions and the red ones denote the ground truth.
	The 2D-3D matches are visualized by projecting the 3D matches of the detected 2D points (shown on the left image) onto the image plane (shown on the right image).
    Perfectly horizontal lines indicate correct 2D-3D matches.
	Reprojection error less than 10 px (in 512 $\times$ 512 images) is colored as green.
	RANSAC is applied to the matches to remove outliers.
	Our method is able to produce a larger quantity of matches compared to baseline HLoc~(\textit{SIFT + NN}). 
	The matches from our method are also more accurate and less noisy compared to the baseline method Ablation (iv)).
	HLoc~(\textit{SPP + SPG}) achieves similar results with our method, but at a ${\sim}10\times$ run-time cost.
	PVNet and Objectron only obtained reasonable results in the first and last examples, respectively. Best viewed in color. 
	}
	\label{fig:qua-allmatch}
	  \vspace{-0.3 cm}
\end{figure*}
\paragraph{Comparison with the Category-level Baseline Objectron}
We compare our method with Objectron~\cite{ahmadyanObjectronLargeScale2020} on all objects in the \textit{Shoe} and \textit{Cup} categories with the metrics used in the original paper 
and present the results in Tab.~\ref{tab:exp_objectron}.
\begin{table}[htb]
\centering
\resizebox{\columnwidth}{!}{%
\setlength\tabcolsep{2pt} %
\begin{tabular}{c|ccc|c|cccc|c} 
 \Xhline{3\arrayrulewidth}
\multicolumn{1}{l|}{Obj. ID}         & \multicolumn{1}{l|}{0415} & \multicolumn{1}{l|}{0475} & \multicolumn{1}{l|}{0476}  & \multicolumn{1}{l|}{\cellcolor[rgb]{0.922,0.918,0.918}Cup}                                       & \multicolumn{1}{l|}{0592} & \multicolumn{1}{l|}{0593} & \multicolumn{1}{l|}{0594} & \multicolumn{1}{l|}{0595}  & \multicolumn{1}{l}{\cellcolor[rgb]{0.922,0.918,0.918}Shoe}                   \\ 
\hline
\multicolumn{10}{c}{Mean pixel error of 2D projection}                                                                                                                                                                                                                                                                                                   \\ 
\hline
Objectron                            & 0.269                     & 0.474                     & 0.483                      & {\cellcolor[rgb]{0.922,0.918,0.918}}0.054                      & 0.189                     & 0.183                     & 0.118                     & 0.124                      & {\cellcolor[rgb]{0.922,0.918,0.918}}0.039  \\
Objectron (S)                        & 0.170                     & 0.340                     & 0.347                      & {\cellcolor[rgb]{0.922,0.918,0.918}}-                          & 0.123                     & 0.115                     & 0.092                     & 0.089                      & {\cellcolor[rgb]{0.922,0.918,0.918}}-      \\
\multicolumn{1}{l|}{Objectron (S+C)} & \multicolumn{1}{l}{0.158} & \multicolumn{1}{l}{0.331} & \multicolumn{1}{l|}{0.342} & {\cellcolor[rgb]{0.922,0.918,0.918}}-                          & \multicolumn{1}{l}{0.103} & \multicolumn{1}{l}{0.098} & \multicolumn{1}{l}{0.084} & \multicolumn{1}{l|}{0.079} & {\cellcolor[rgb]{0.922,0.918,0.918}}-      \\
Ours                                 & \textbf{0.047}            & \textbf{0.022}            & \textbf{0.013}             & {\cellcolor[rgb]{0.922,0.918,0.918}}-                          & \textbf{0.089}            & \textbf{0.016}            & \textbf{0.026}            & \textbf{0.012}             & {\cellcolor[rgb]{0.922,0.918,0.918}}-      \\ 
\hline
\multicolumn{10}{c}{Average precision at 15 $\degree$ Azimuth error}                                                                                                                                                                                                                                                                                     \\ 
\hline
Objectron                            & 0.364                     & 0.131                     & 0.217                      & \multicolumn{1}{l|}{{\cellcolor[rgb]{0.922,0.918,0.918}}0.644} & 0.677                     & 0.733                     & 0.774                     & 0.945                      & {\cellcolor[rgb]{0.922,0.918,0.918}}0.586  \\
Ours                                 & \textbf{1.0}              & \textbf{1.0}              & \textbf{1.0}               & {\cellcolor[rgb]{0.922,0.918,0.918}}-                          & \textbf{0.855}            & \textbf{0.998}            & \textbf{0.984}            & \textbf{1.0}               & {\cellcolor[rgb]{0.922,0.918,0.918}}-      \\ 
\hline
\multicolumn{10}{c}{Average precision at 10 $\degree$ Elevation error}                                                                                                                                                                                                                                                                                   \\ 
\hline
Objectron                            & 0.707                     & 0.906                     & 0.821                      & {\cellcolor[rgb]{0.922,0.918,0.918}}0.837                      & 0.794                     & 0.842                     & 0.622                     & 0.995                      & {\cellcolor[rgb]{0.922,0.918,0.918}}0.754  \\
Ours                                 & \textbf{1.0}              & \textbf{1.0}              & \textbf{1.0}               & {\cellcolor[rgb]{0.922,0.918,0.918}}-                          & \textbf{0.831}            & \textbf{0.996}            & \textbf{0.973}            & \textbf{1.0}               & {\cellcolor[rgb]{0.922,0.918,0.918}}-      \\
 \Xhline{3\arrayrulewidth}
\end{tabular}
}
\caption{\textbf{Comparison with the \textit{category-level} baseline.}
Our method is compared with Objectron\cite{ahmadyanObjectronLargeScale2020}
 with auxiliary scale adjustment \textit{(S)} and center alignment  \textit{(S+C)}. The category-level evaluation results reported in the original paper are provided in grey background below the name of the category. }
  \vspace{-0.3 cm}
\label{tab:exp_objectron}
\end{table}
For mean pixel error of 2D projection, the results of Objectron on our dataset are far from the reported results for the two categories on Objectron dataset.
This is because of the deviations in ground-truth annotations between the Objectron dataset and our dataset.
For a fair comparison, we further apply scaling and center alignment operations to the predictions of Objectron to alleviate this gap and 
provided results respectively as \textit{Objectron (S)} and \textit{Objectron (S+C)} in Tab.~\ref{tab:exp_objectron}.
Although the performances of Objectron do get boosted and are comparable with the reported results in the original paper, our method surpasses it by a large margin.
Our method outperforms Objectron evidently in the average precision of azimuth error and elevation error, 
especially for the objects of \textit{Cup} category where the shape and appearance may vary significantly between instances.
These experiments illustrate the limited generalization ability of category-level methods to new object instances.

\paragraph{Runtime Analysis}
We report the runtimes of our visual localization baselines and our method in Tab. \ref{tab:exp_visloc}.
The runtime consist of feature extraction for the query image with SuperPoint and the 2D-3D matching process without 2D detection and PnP.
Our method runs ${\sim}10\times$ faster than HLoc (\textit{SPP + SPG}).
All the experiments are conducted on an NVIDIA TITAN RTX GPU.

\subsection{Ablation Studies}
\label{sub_sec:ablation}
In this section. we conduct several ablation experiments by substituting GATs with simpler counterparts of the feature aggregation and matching modules. 
All the results for our ablation studies are presented in Tab.~\ref{tab:exp_abl_single}.

\paragraph{Effectiveness of the Aggregation-Attention Layer}
We validate the effectiveness of the proposed aggregation-attention layers by substituting the corresponding aggregation layers in GATs by the averaging operation and report the result in Tab.~\ref{tab:exp_abl_single} as (i).
Notice the 2D-3D matching is still based on a GNN with self- and cross-attention layers, which is similar to SuperGlue.
Without the aggregation-attention layers, the results dropped significantly for large and medium objects, which indicates the effectiveness of aggregation-attention layers.
\begin{table}[htb]
\centering
\resizebox{\columnwidth}{!}{%
\setlength\tabcolsep{2pt} %
\begin{tabular}{cccc||c|c|c} 
 \Xhline{3\arrayrulewidth}
\multicolumn{4}{c||}{Components}              & \multicolumn{1}{c}{\multirow{2}{*}{Large Objects}} & \multicolumn{1}{c}{\multirow{2}{*}{Medium Objects}} & \multirow{2}{*}{Small Objects}  \\ 
\cline{1-4}
              & 2D Feat. & Aggr.   & Matching & \multicolumn{1}{c}{}                               & \multicolumn{1}{c}{}                                &                                 \\ 
\hline
\textit{Ours} & SPP      & GATs    & GNN      & \textbf{0.471}                                     & \textbf{0.629}                                      & \textbf{0.405}                  \\ 
\hline
i             & SPP      & Avg.    & GNN      & 0.449                                              & 0.602                                               & 0.390                           \\
ii            & SPP      & Avg.    & NN       & 0.426                                              & 0.570                                               & 0.367                           \\
iii           & SPP      & K-Means & NN       & 0.431                                              & 0.595                                       & 0.379                   \\
iv            & SIFT     & Avg.    & NN       & 0.355                                              & 0.470                                               & 0.256                           \\
v             & SIFT     & K-Means & NN       & 0.369                                              & 0.450                                               & 0.281                           \\
 \Xhline{3\arrayrulewidth}
\end{tabular}
}
\caption{\textbf{Ablation studies.}
Different alternatives for components in the proposed method are compared with the \textit{1cm-1deg} metric.
SPP stands for SuperPoint and 
NN stands for Nearest Neighbor.
} 
  \vspace{-0.3 cm}
\label{tab:exp_abl_single}
\end{table}
The simple averaging operation cannot adaptively select relevant information from different viewpoints according to different query features.

\paragraph{Other variants with 2D-3D NN Matching}
To provide more comparisons with traditional pipelines~\cite{skrypnykSceneModellingRecognition2004,Collet2009ObjectRA,tangTexturedObjectRecognition2012} that estimate object pose with local features and 2D-3D matching, 
we also experimented with variants of our method based on different local features, feature aggregration methods and matchers for 2D-3D matching. The results are 
reported in Tab.~\ref{tab:exp_abl_single} as (ii - v).
(ii - v) are still unable to produce comparable results with our approach. 
Compared with (ii) and (iv) that use averaging for feature aggregation, our method consistently outperforms them by a significant margin.
Similar to the analysis for (i), simply averaging the features from different viewpoints loses view-dependent information.
For (iii) and (v), 
substituting averaging with K-Means clustering could provide richer 3D features but the results are still not comparable with ours.

\paragraph{Qualitative Comparisons}
We provide some qualitative results to compare our method with baseline methods in Fig.~\ref{fig:qua-allmatch}.
Please read the caption for details.
\section{Conclusion}
In this paper, we propose \shortname for one-shot object pose estimation.
Unlike existing instance-level or category-level methods, OnePose does not rely on CAD models and can handle objects in arbitrary categories without instance- or category-specific network training.
Compared with localization-based baseline methods, instance-level baseline method PVNet and category-level baseline method Objectron, \shortname achieves better pose estimation accuracy and faster inference speed.
We also believe that our revisit to the localization-based setting (i.e., one-shot object pose estimation) is more practical for AR and valuable to the community.  %

\paragraph{Limitations}
The limitations of our method come with the nature of relying on local feature matching for pose estimation.
Our method may fail when applied to textureless objects. 
Although being enhanced by attention mechanisms, 
our method still has difficulty to handle extreme change of scales
 between images in the video scan and the testing sequences.

\paragraph{Acknowledgements}
The authors would like to acknowledge the support from the National Key Research and Development Program of China (No. 2020AAA0108901), NSFC (No. 62172364), and the ZJU-SenseTime Joint Lab of 3D Vision.

{
    \clearpage
    \small
    \bibliographystyle{ieee_fullname}
        \bibliography{macros, one_pose_clean, ext_db_clean}

\begin{thebibliography}{10}\itemsep=-1pt

\bibitem{arcore}
{ARC}ore.
\newblock \url{https://developers.google.com/ar}.

\bibitem{arkit}
{ARK}it.
\newblock \url{https://developer.apple.com/augmented-reality/}.

\bibitem{UltralyticsYolov52021}
Ultralytics/yolov5.
\newblock \url{https://github.com/ultralytics/yolov5}, 2021.

\bibitem{ahmadyanObjectronLargeScale2020}
Adel Ahmadyan, Liangkai Zhang, Jianing Wei, Artsiom Ablavatski, and Matthias
  Grundmann.
\newblock Objectron: A large scale dataset of object-centric videos in the wild
  with pose annotations.
\newblock {\em CVPR}, 2021.

\bibitem{fleet_learning_2014}
Eric Brachmann, Alexander Krull, Frank Michel, Stefan Gumhold, Jamie Shotton,
  and Carsten Rother.
\newblock Learning {6D} object pose estimation using {3D} object coordinates.
\newblock In {\em ECCV}. 2014.

\bibitem{Brachmann_2016_CVPR}
Eric Brachmann, Frank Michel, Alexander Krull, Michael~Ying Yang, Stefan
  Gumhold, and Carsten Rother.
\newblock Uncertainty-driven {6D} pose estimation of objects and scenes from a
  single {RGB} image.
\newblock In {\em CVPR}, 2016.

\bibitem{caiReconstructLocallyLocalize}
Ming Cai and Ian Reid.
\newblock Reconstruct locally, localize globally: {A} model free method for
  object pose estimation.
\newblock In {\em CVPR}, 2020.

\bibitem{chen_category_nodate}
Xu Chen, Zijian Dong, Jie Song, Andreas Geiger, and Otmar Hilliges.
\newblock Category level object pose estimation via neural
  analysis-by-synthesis.
\newblock {\em ECCV}, 2020.

\bibitem{Collet2009ObjectRA}
Alvaro Collet, Dmitry Berenson, Siddhartha~S. Srinivasa, and Dave Ferguson.
\newblock Object recognition and full pose registration from a single image for
  robotic manipulation.
\newblock {\em ICRA}, 2009.

\bibitem{detoneSuperPointSelfSupervisedInterest2018}
Daniel DeTone, Tomasz Malisiewicz, and Andrew Rabinovich.
\newblock {{SuperPoint}}: Self-supervised interest point detection and
  description.
\newblock {\em CVPRW}, 2017.

\bibitem{detoneGeometricDeepSLAM2017}
Daniel DeTone, Tomasz Malisiewicz, and Andrew Rabinovich.
\newblock Toward geometric deep slam.
\newblock {\em arXiv:1707.07410 [cs]}, 2017.
\newblock arXiv: 1707.07410.

\bibitem{dusmanuD2NetTrainableCNN2019}
Mihai Dusmanu, Ignacio Rocco, Tomas Pajdla, Marc Pollefeys, Josef Sivic,
  Akihiko Torii, and Torsten Sattler.
\newblock D2-{{Net}}: A trainable cnn for joint detection and description of
  local features.
\newblock {\em CVPR}, 2019.

\bibitem{hutchison_model_2013}
Stefan Hinterstoi{\ss}er, Vincent Lepetit, Slobodan Ilic, Stefan Holzer,
  Gary~R. Bradski, Kurt Konolige, and Nassir Navab.
\newblock Model based training, detection and pose estimation of texture-less
  3d objects in heavily cluttered scenes.
\newblock In {\em ACCV}. 2012.

\bibitem{katharopoulos_et_al_2020}
Angelos Katharopoulos, Apoorv Vyas, Nikolaos Pappas, and Fran{\c{c}}ois
  Fleuret.
\newblock Transformers are {RNNs}: Fast autoregressive transformers with linear
  attention.
\newblock In {\em ICML}, Proceedings of Machine Learning Research, 2020.

\bibitem{kehl_ssd-6d_2017}
Wadim Kehl, Fabian Manhardt, Federico Tombari, Slobodan Ilic, and Nassir Navab.
\newblock {SSD-6D:} making rgb-based 3d detection and 6d pose estimation great
  again.
\newblock In {\em ICCV}, 2017.

\bibitem{labbeCosyPoseConsistentMultiview2020}
Yann Labb{\'e}, Justin Carpentier, Mathieu Aubry, and Josef Sivic.
\newblock {{CosyPose}}: Consistent multi-view multi-object {{6D}} pose
  estimation.
\newblock {\em ECCV}, 2020.

\bibitem{lee_category-level_2021}
Taeyeop Lee, Byeong-Uk Lee, Myungchul Kim, and In~So Kweon.
\newblock Category-{Level} metric scale object shape and pose estimation.
\newblock {\em IEEE Robotics and Automation Letters}, 2021.

\bibitem{ferrari_deepim:_2018}
Yi Li, Gu Wang, Xiangyang Ji, Yu Xiang, and Dieter Fox.
\newblock {DeepIM}: Deep iterative matching for 6d pose estimation.
\newblock In {\em ECCV}. 2018.

\bibitem{linFocalLossDense2018}
Tsung{-}Yi Lin, Priya Goyal, Ross~B. Girshick, Kaiming He, and Piotr
  Doll{\'{a}}r.
\newblock Focal loss for dense object detection.
\newblock In {\em ICCV}, 2017.

\bibitem{loweDistinctiveImageFeatures2004}
David~G. Lowe.
\newblock Distinctive image features from scale-invariant keypoints.
\newblock {\em IJCV}, 2004.

\bibitem{martinezMOPEDScalableLow2010}
Manuel Martinez, Alvaro Collet, and Siddhartha~S Srinivasa.
\newblock {{MOPED}}: A scalable and low latency object recognition and pose
  estimation system.
\newblock In {\em ICRA}, 2010.

\bibitem{oberweger_making_2018-1}
Markus Oberweger, Mahdi Rad, and Vincent Lepetit.
\newblock Making deep heatmaps robust to partial occlusions for {3D} object
  pose estimation.
\newblock {\em ECCV}, 2018.

\bibitem{parkLatentFusionEndtoEndDifferentiable2019}
Keunhong Park, Arsalan Mousavian, Yu Xiang, and Dieter Fox.
\newblock {LatentFusion}: End-to-end differentiable reconstruction and
  rendering for unseen object pose estimation.
\newblock In {\em CVPR}, 2020.

\bibitem{park_pix2pose:_2019}
Kiru Park, Timothy Patten, and Markus Vincze.
\newblock {Pix2Pose}: Pixel-wise coordinate regression of objects for 6d pose
  estimation.
\newblock In {\em ICCV}, 2019.

\bibitem{pavlakos6DoFObjectPose2017}
Georgios Pavlakos, Xiaowei Zhou, Aaron Chan, Konstantinos~G. Derpanis, and
  Kostas Daniilidis.
\newblock {6-DoF} object pose from semantic keypoints.
\newblock {\em ICRA}, 2017.

\bibitem{pengPVNetPixelwiseVoting2018}
Sida Peng, Yuan Liu, Qixing Huang, Xiaowei Zhou, and Hujun Bao.
\newblock {PVNet}: Pixel-wise voting network for 6dof pose estimation.
\newblock In {\em CVPR}, 2019.

\bibitem{peng2020pvnet}
Sida Peng, Xiaowei Zhou, Yuan Liu, Haotong Lin, Qixing Huang, and Hujun Bao.
\newblock {PVNet}: pixel-wise voting network for 6dof object pose estimation.
\newblock {\em T-PAMI}, 2020.

\bibitem{qiuTracking3DMotion2019}
Kejie Qiu, Tong Qin, Wenliang Gao, and Shaojie Shen.
\newblock Tracking 3-{D} motion of dynamic objects using monocular
  visual-inertial sensing.
\newblock {\em IEEE Transactions on Robotics}, (4), 2019.

\bibitem{leonardis_machine_2006}
Edward Rosten and Tom Drummond.
\newblock Machine learning for high-speed corner detection.
\newblock In {\em ECCV}. 2006.

\bibitem{rubleeORBEfficientAlternative2011}
Ethan Rublee, Vincent Rabaud, Kurt Konolige, and Gary~R. Bradski.
\newblock {ORB:} an efficient alternative to {SIFT} or {SURF}.
\newblock In {\em ICCV}, 2011.

\bibitem{sarlinCoarseFineRobust2019}
Paul{-}Edouard Sarlin, Cesar Cadena, Roland Siegwart, and Marcin Dymczyk.
\newblock From coarse to fine: Robust hierarchical localization at large scale.
\newblock In {\em CVPR}, 2019.

\bibitem{sarlinSuperGlueLearningFeature2020}
Paul{-}Edouard Sarlin, Daniel DeTone, Tomasz Malisiewicz, and Andrew
  Rabinovich.
\newblock {SuperGlue}: Learning feature matching with graph neural networks.
\newblock In {\em ICCV}, 2020.

\bibitem{schonbergerStructurefromMotionRevisited2016}
Johannes~L. Sch{\"{o}}nberger and Jan{-}Michael Frahm.
\newblock Structure-from-motion revisited.
\newblock In {\em CVPR}, 2016.

\bibitem{shotton_scene_2013}
Jamie Shotton, Ben Glocker, Christopher Zach, Shahram Izadi, Antonio Criminisi,
  and Andrew~W. Fitzgibbon.
\newblock Scene coordinate regression forests for camera relocalization in
  {RGB-D} images.
\newblock In {\em CVPR}, 2013.

\bibitem{skrypnykSceneModellingRecognition2004}
I. Skrypnyk and D.G. Lowe.
\newblock Scene modelling.
\newblock In {\em ISMAR}.

\bibitem{sunLoFTRDetectorFreeLocal2021}
Jiaming Sun, Zehong Shen, Yuang Wang, Hujun Bao, and Xiaowei Zhou.
\newblock {{LoFTR}}: Detector-free local feature matching with transformers.
\newblock {\em CVPR}, 2021.

\bibitem{tangTexturedObjectRecognition2012}
Jie Tang, Stephen Miller, Arjun Singh, and P. Abbeel.
\newblock A textured object recognition pipeline for color and depth image
  data.
\newblock {\em ICRA}, 2012.

\bibitem{tian_shape_2020}
Meng Tian, Marcelo~H. Ang~Jr, and Gim~Hee Lee.
\newblock Shape prior deformation for categorical {6D} object pose and size
  estimation.
\newblock {\em ECCV}, 2020.

\bibitem{tyszkiewiczDISKLearningLocal2020}
Michal~J. Tyszkiewicz, Pascal Fua, and Eduard Trulls.
\newblock {DISK:} learning local features with policy gradient.
\newblock In {\em NeurIPS}, 2020.

\bibitem{velickovicGraphAttentionNetworks2018}
Petar Velickovic, Guillem Cucurull, Arantxa Casanova, Adriana Romero, Pietro
  Li{\`{o}}, and Yoshua Bengio.
\newblock Graph attention networks.
\newblock In {\em ICLR}, 2018.

\bibitem{vyas_et_al_2020}
Apoorv Vyas, Angelos Katharopoulos, and Fran{\c{c}}ois Fleuret.
\newblock Fast transformers with clustered attention.
\newblock In {\em NeurIPS}, 2020.

\bibitem{wang_gdr-net_2021}
Gu Wang, Fabian Manhardt, Federico Tombari, and Xiangyang Ji.
\newblock {GDR}-{Net}: Geometry-guided direct regression network for monocular
  {6D} object pose estimation.
\newblock In {\em CVPR}, 2021.

\bibitem{wangNormalizedObjectCoordinate2019}
He Wang, Srinath Sridhar, Jingwei Huang, Julien Valentin, Shuran Song, and
  Leonidas~J. Guibas.
\newblock Normalized object coordinate space for category-level 6d object pose
  and size estimation.
\newblock In {\em CVPR}, 2019.

\bibitem{wang_category-level_2021}
Jiaze Wang, Kai Chen, and Qi Dou.
\newblock Category-level {6D} object pose estimation via cascaded relation and
  recurrent reconstruction networks.
\newblock {\em IROS}, 2021.

\bibitem{wenBundleTrack6DPose2021}
B Wen and Kostas~E Bekris.
\newblock {BundleTrack}: 6d pose tracking for novel objects without instance or
  category-level 3d models.
\newblock {\em ICRA}, 2021.

\bibitem{xiangPoseCNNConvolutionalNeural2017}
Yu Xiang, Tanner Schmidt, Venkatraman Narayanan, and Dieter Fox.
\newblock {PoseCNN}: A convolutional neural network for 6d object pose
  estimation in cluttered scenes.
\newblock {\em RSS}, 2018.

\end{thebibliography}
}

\end{document}